\documentclass{bmvc2k}

\title{Scalable Frame Sampling for Video Classification: A Semi-Optimal Policy Approach with Reduced Search Space}

\addauthor{Junho Lee}{joon2003@snu.ac.kr}{1}
\addauthor{Jeongwoo Shin}{swswss@snu.ac.kr}{1}
\addauthor{Seung Woo Ko}{seungwoo.ko@lgresearch.ai}{1,2}
\addauthor{Seongsu Ha}{mars@twelvelabs.io}{1,3}
\addauthor{Joonseok Lee$^*$}{joonseok@snu.ac.kr}{1,4}

\addinstitution{
 Seoul National University \\
 Seoul, Korea
}
\addinstitution{
 LG AI Research \\
 Seoul, Korea
}
\addinstitution{
 Twelve Labs \\
 Seoul, Korea
}
\addinstitution{
 Google Research \\
 Mountain View, CA, USA\\
 \hspace{-0.25cm} $^*$Corresponding author
}

\runninghead{Lee et al.}{Scalable Frame Sampling for Video Classification}

\usepackage{caption}
\usepackage{sidecap}
\usepackage{xcolor}  
\usepackage{colortbl}
\usepackage{graphicx}
\usepackage{dsfont}
\usepackage{amsmath,amssymb}
\usepackage{bm}

\definecolor{maroon}{cmyk}{0,0.87,0.68,0.32}
\definecolor{LightGray}{gray}{0.75}

\definecolor{ashgrey}{rgb}{0.7, 0.75, 0.71}
\definecolor{lavendergray}{rgb}{0.77, 0.76, 0.82}

\newcommand\rone[1]{\textbf{\textcolor{orange}{z2x8}}}
\newcommand\rtwo[1]{\textbf{\textcolor{blue}{NaYm}}}
\newcommand\rthree[1]{\textbf{\textcolor{cyan}{EsVC}}}

\AtEndPreamble{
    \usepackage[capitalize]{cleveref}
    \crefname{section}{Sec.}{Secs.}
    \Crefname{section}{Section}{Sections}
    \crefname{table}{Tab.}{Tabs.}
    \Crefname{table}{Table}{Tables}
}
\begin{document}

\maketitle

    \begin{abstract}
Given a video with $T$ frames, frame sampling is a task to select $N \ll T$ frames, so as to maximize the performance of a fixed video classifier.
Not just brute-force search, but most existing methods suffer from its vast search space of $\binom{T}{N}$, especially when $N$ gets large.
To address this challenge, we introduce a novel perspective of reducing the search space from $O(T^N)$ to $O(T)$.
Instead of exploring the entire $O(T^N)$ space, our proposed semi-optimal policy selects the top $N$ frames based on the independently estimated value of each frame using per-frame confidence, significantly reducing the computational complexity.
We verify that our semi-optimal policy can efficiently approximate the optimal policy, particularly under practical settings. Additionally, through extensive experiments on various datasets and model architectures, we demonstrate that learning our semi-optimal policy ensures stable and high performance regardless of the size of $N$ and $T$.
Our code is available at https://github.com/isno0907/sosampler.


\end{abstract}

\section{Introduction}
\label{sec:intro}
As video platforms continue to grow explosively, efficient and scalable video understanding becomes increasingly important.
With remarkable advances in deep learning, action recognition and video understanding have also made tremendous progress, from 2D~\cite{ng2015snippets,fernando2016rankpooling,zhou2018trn, simonyan2014twostream,feichtenhofer2016twostreamfusion,lee2018cdml,lee2020gcml} and 3D CNN models \cite{tran2015c3d,carreira2017i3d,qiu2017p3d, tran2018r2+1d,feichtenhofer2019slowfast,tran2019csn} to Transformers \cite{arnab2021vivit,bertasius2021timesformer,liu2022videoswin}.
Despite these advances, video understanding models are still often overwhelmed by high storage and computational cost.
To mitigate this, lighter models with less parameters and computational cost \cite{zolfaghari2018eco,kopuklu2019resource,fan2019moreisless,feichtenhofer2020x3d,verelst2020dynamic,piergiovanni2022tvn} have been proposed.

Focusing mostly on the modeling aspect, however, these works have not touched the most basic underlying condition that digital videos are encoded as a temporal sequence of regularly sampled frames, where each frame is a 2D spatial matrix of regularly sampled pixels, regardless of the content.
These regularly sampled pixels are often highly redundant, especially when the frame rate is high.
Ideally, it will be more efficient to sample frames proportional to the amount of information, removing redundancy among them as much as possible.
Given that most existing video models are still built on a constant frame sampling rate for all videos, there are further room for improvement in terms of efficiency.

In this paper, we consider the frame sampling problem. For a video with $T$ frames, we propose a frame sampler that selects $N \ll T$ frames, which are then used by a video model pre-trained for a specific downstream task (e.g., classification).
The goal of this task is to train a sampler that finds the best combination of $N$ frames out of the given $T$ frames, letting the pre-trained model (classifier) to achieve the highest downstream task performance.

To achieve this, the sampler learns a \emph{policy} to select $N$ frames out of $T$ candidates.
The ideal policy would select the best $N$ frames that lead to the highest downstream task performance, and we call this \emph{optimal policy} ($\pi_o$).
A naive approach would collect the best combinations as training data and train the sampler in a supervised manner.
However, finding the best combination is an NP-hard combinatorial optimization problem, as it requires comparing all $\binom{T}{N}$ possible combinations to find the solution.
Exhaustively searching this $O(T^N)$ space is practically infeasible when $N$ and $T$ grow.

Several previous works~\cite{wu2019adaframe,wang2022adafocus,lin2022ocsampler,li2021ada2d,wu2019marl} have applied reinforcement learning to search this space, setting up the sampler as an agent and the classifier as the environment and training the sampler with a reward function designed to find $\pi_o$.
While they have shown promising results for small $N$ and $T$, the underlying search space of $O(T^N)$ still remains the same, posing challenges for large $N$ and $T$.




Considering the purpose of the frame sampling task, however, it is more important for the sampler to operate effectively on a long video with a large $N$ and $T$.
Unlike the previous works that operate directly within the $O(T^N)$ space, we take a step back to find a way to reduce this search space itself.
The fundamental reason for this exponential growth of search space is that the value of a selected frame depends on those of other selected frames and thus we need to consider the joint distribution of the frame values.

What if, however, the value of a frame can be fairly determined independently from each other?
We would be able to score $T$ frames independently and simply to take the top $N$ frames.
In this paper, we argue that, based on experimental results, frames can be reasonably assessed independently under most practical conditions where frame sampling is relevant—namely, when the video is not excessively short and the frame rate is not extremely high. We observe that a policy based on evaluating the value of each frame can reasonably approximate an optimal sampling policy.

Based on this observation, we propose a \textit{semi-optimal policy} ($\pi_s$) that selects the top $N$ frames based on estimated value of each frame using per-frame confidence.
This approach significantly reduces the search space to $O(T)$.
We empirically show on multiple datasets that $\pi_s$ is a policy that reasonably approximates $\pi_o$, and demonstrate that the state-of-the-art sampler~\cite{lin2022ocsampler} achieves more stable and superior performance when it learns $\pi_s$ instead of $\pi_o$, not only with small but also large $N$ and $T$.

Our contributions are summarized as follows:
\begin{itemize}
    \setlength{\itemsep}{0pt}
    \setlength{\parskip}{0pt}
    \item We introduce a novel perspective of \textit{reducing the search space from $O(T^N)$ to $O(T)$} in frame sampling for video classification, proposing the semi-optimal policy.
    \item {Through various analyses, we demonstrate that the semi-optimal policy approximates the optimal policy \textit{under the most practical settings}.}
    \item {From extensive experiments, we show that a sampler learning the semi-optimal policy achieves stable and high performance across both small and large values of $N$ and $T$.}
\end{itemize}

\section{Related Work}
\label{sec:related}

\noindent
\textbf{Video Recognition.}
2D and 3D convolutional neural networks (CNNs) have been widely used for image and video classification.
2D-CNN encodes each frame individually, then aggregates the temporal dynamics by feature pooling~\cite{ng2015snippets,lee2018yt8m2,hwang2019video}, rank pooling~\cite{fernando2016rankpooling}, or relation network~\cite{zhou2018trn}.
Two-stream methods take both spatial stream from CNN and temporal stream from optical flow as input at various levels of fusion~\cite{simonyan2014twostream,feichtenhofer2016twostreamfusion}.
3D-CNN approaches apply convolution operations on both spatial and temporal dimensions simultaneously, pioneered by C3D~\cite{tran2015c3d}.
I3D~\cite{carreira2017i3d} combines the 3D convolution with two-stream approaches.
P3D~\cite{qiu2017p3d} and R(2+1)D~\cite{tran2018r2+1d} factorizes the spatial and temporal dimensions in the 3D convolution.
SlowFast~\cite{feichtenhofer2019slowfast} model consists of two different pathways of focusing more on temporal and spatial information, respectively.
CSN~\cite{tran2019csn} shows that separating the channel interaction with spatio-temporal interaction leads to lower computational cost and better regularization.

Recently, Transformers~\cite{vaswani2017attention} have been applied to video classification and action recognition.
Due to the high computational cost for processing all patches in the video, most video Transformer models factorize spatial and temporal attentions~\cite{arnab2021vivit,bertasius2021timesformer} or utilize the inductive bias of locality in videos \cite{liu2022videoswin} for efficient handling of videos.

\vspace{0.1cm} \noindent
\textbf{Efficient Video Recognition.}
One of the biggest challenges of video classification is the computational cost stemming from the vast number of frames.
One remedy is to lighten the model architecture itself~\cite{zolfaghari2018eco,kopuklu2019resource,fan2019moreisless,feichtenhofer2020x3d,verelst2020dynamic,piergiovanni2022tvn}.
Another direction is algorithmic improvement for better efficiency.
AdaFrame~\cite{wu2019adaframe}
learns from both the current frame and the global context to make decisions on where to look next in the video.
AdaFocus series~\cite{wang2022adafocus, wang2021adafocus} use a lightweight global CNN to guide the policy network to extract the most useful image crop to classify the video.
On-the-fly gating techniques are introduced to reduce computation by only extracting finer features at selected frames~\cite{wu2019liteeval}, determining an early exiting point~\cite{ghodrati2021frameexit}, or deciding which frames in the video to fully process~\cite{seon2023stoporforward}. ListenToLook~\cite{gao2020listentolook} utilizes only a single frame and audio stream to determine the full video descriptor for the clip.

An alternative is to sample frames to process, instead of feeding all of them to the classification model, vastly reducing the computational cost.
ARNet~\cite{meng2020arnet} and VideoIQ~\cite{sun2021videoiq} pre-train multiple classifiers and train a sampler to determine which classifier to process each frame. MARL~\cite{wu2019marl} utilizes multiple agents that adjust the sampling location based on local and historical context. OCSampler~\cite{lin2022ocsampler} simply selects the salient frames, while Ada2D~\cite{li2021ada2d} determines which frames will be used in 2D or 3D.
Ours falls into this category, particularly aiming to effectively sample $N$ frames for a large $T$.

\section{Problem Formulation}
\label{sec:problem}

\noindent
\textbf{Sampling Scenario.}
We assume an offline environment, where we are given $T$ candidate frames $\bm{v} \in \mathbb{R}^{T \times 3 \times H \times W}$ from a video, with $H$ and $W$ indicating the height and width of the frames, respectively.
A frozen pretrained classifier $f_c: \mathbb{R}^{N \times 3 \times H \times W} \rightarrow [0,1]^{C}$ is given, where $N$ is the number of input frames and $C$ is the number of classes.
In this context, our final objective is to train a sampler that finds the optimal $N$ frames out of the given $T$ frames that make the classifier to best perform on the downstream task.

\vspace{0.1cm}
\noindent\textbf{Optimal Policy $\bm{\pi_o}$.}
Given $T$ candidate frames, the optimal policy $\pi_o$ chooses $N$ frames that maximize the confidence of the classifier $f_c$ on the true label $y \in [1, ..., C]$, among all possible combinations of the candidate frames. We define the frames selected by $\pi_o$ as optimal set.


\vspace{0.1cm}
\noindent\textbf{Challenges.}
The most straightforward approach to learn $\pi_o$ would be to train a sampler in a supervised manner to select the optimal set of frames.
However, finding the optimal set of frames is computationally prohibitive, requiring to explore a search space of $O(T^N)$.
Even for moderately large $N$ and $T$, the search space quickly grows, making it challenging to train the sampler directly.

To explore this space, previous works adopt reinforcement learning, where a network sampler (agent) is trained based on rewards given by the classifier (environment) for the selected frames (action) according to the policy.
Although these approaches have been effective for small $N$ and $T$, they are not scalable for growing $N$ and $T$.
To mitigate the exponential time complexity caused by this exponentially growing search space, we propose an effective approach in most practical settings.


\section{Method}
\label{sec:method}
\subsection{Semi-Optimal Policy}
\label{subsec:semi}
The main challenge of the frame sampling problem lies in the fact that the importance of a frame is influenced by other selected frames, necessitating to explore the interactions in the exponentially growing search space to $N$ and $T$.
Conversely, we recognize that the problem would be much simpler if: 1) the importance of each frame were (roughly) independent from other frames, and 2) we could reasonably approximate the importance of each frame.
This would allow us to score $T$ frames individually and select the top $N$ of them.

In this section, we show that the frames are indeed close to independence when it comes to a practical setting of the frame sampling task -- where the video is not trivially short and the frame rate is relatively low.
The opposite case, a short snippet with high frame rate, would not need a sophisticated frame sampling method anyway,
since most frames would be highly redundant.
From this observation,
we introduce a new sampling policy that effectively approximates the optimal policy while significantly reducing the space complexity.


\vspace{0.1cm} \noindent
\textbf{Independence between Frames.}
We claim that frames influence each other's importance mainly due to the overlapping information between them.
That is, even if a frame contains important information, its importance will be diminished if the same information is redundantly provided by others.
Conversely, the importance of a frame increases if it contains accurate and unique information; in other words, if it provides distinct information that is not covered by other frames.
Thus, if a candidate frame is sufficiently dissimilar from others, we would be able to independently score it without concerning redundancy.

To check applicability of this idea, we measure the similarity between adjacent frames at various frame rates.
The similarity can be computed in several resolutions; as coarse as the probability distribution over the class labels, or as fine-grained as the pixel-level.
Both extremes are less ideal, since they do not convey high-level semantics of each frame.
Thus, we estimate the relevance between two sampled frames $i$ and $j$ by applying a Gaussian smoothing kernel to their visual embeddings, denoted by $\mathbf{x}_i$ and $\mathbf{x}_j$:
\begin{equation}
  \mathcal{I}(\mathbf{x}_i, \mathbf{x}_j) = 
  \frac{1}{Z} e^{\left\{ -\frac{1}{2} (\mathbf{x}_i - \mathbf{x}_j)^\top \Sigma^{-1} (\mathbf{x}_i - \mathbf{x}_j) \right\}}, 
\end{equation}
where $\Sigma$ determines the kernel bandwidth and $Z = 1/|\sqrt{2\pi \Sigma}|$ is the normalizer.



\begin{figure}[t]
  \centering
  \includegraphics[width=\textwidth]{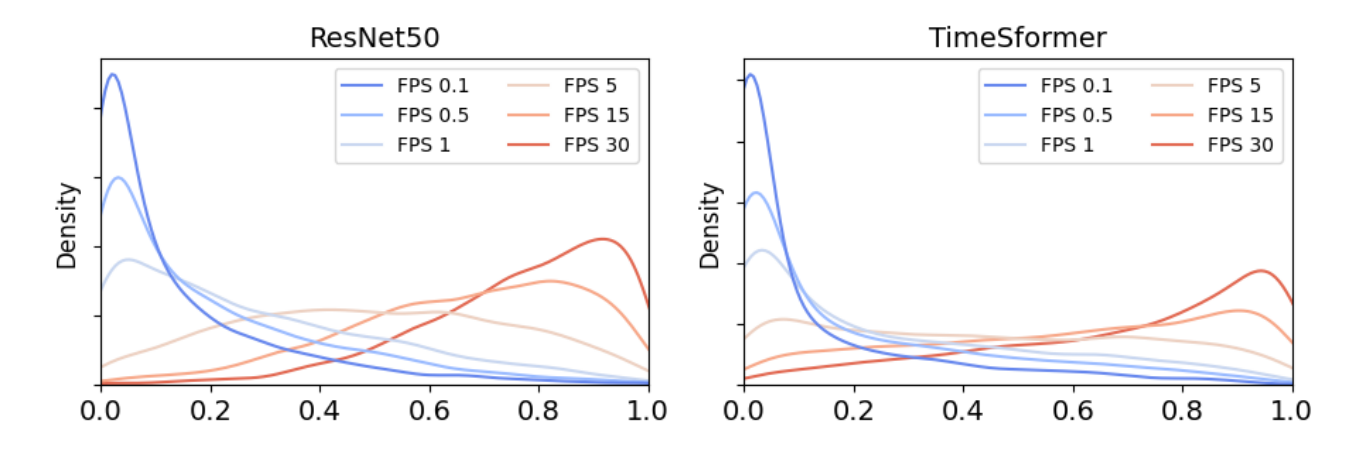}
  \vspace{-0.2cm}
  \caption{\textbf{Distribution of $\mathcal{I}(\mathbf{x}_i, \mathbf{x}_j)$} with kernel density estimate.}
  \label{fig:mutual_info}
\end{figure}

\cref{fig:mutual_info} illustrates the estimated relevance
between two consecutive frames at various frame rates on ActivityNet-v1.3, using ResNet50 and TimeSformer features.
We observe with both features that higher frame rates lead to higher redundancy between frames.
Although this tendency is expected considering the nature of videos, what we need to focus is \emph{where is the proper frame rate threshold} that we can safely assume independence between frames.
From the plots, we conclude that we can reliably assume independence up to 1 fps, while 5 fps is on the borderline.
Under a practical setting for frame sampling, we claim that 1 fps is still reasonable, since higher frame rates would be prohibitively expensive for long videos; \emph{e.g.}, for a 5-min-long video, 5 fps already piles up 1,500 candidate frames.
In short, we can reasonably assume independence among frames when the target video is not trivially short, since we would only feed regularly sampled frames at lower frame rate due to the computational overhead.
\begin{figure}
    \centering
    \begin{minipage}{0.48\textwidth}
        \centering
        \includegraphics[width=0.9\textwidth]{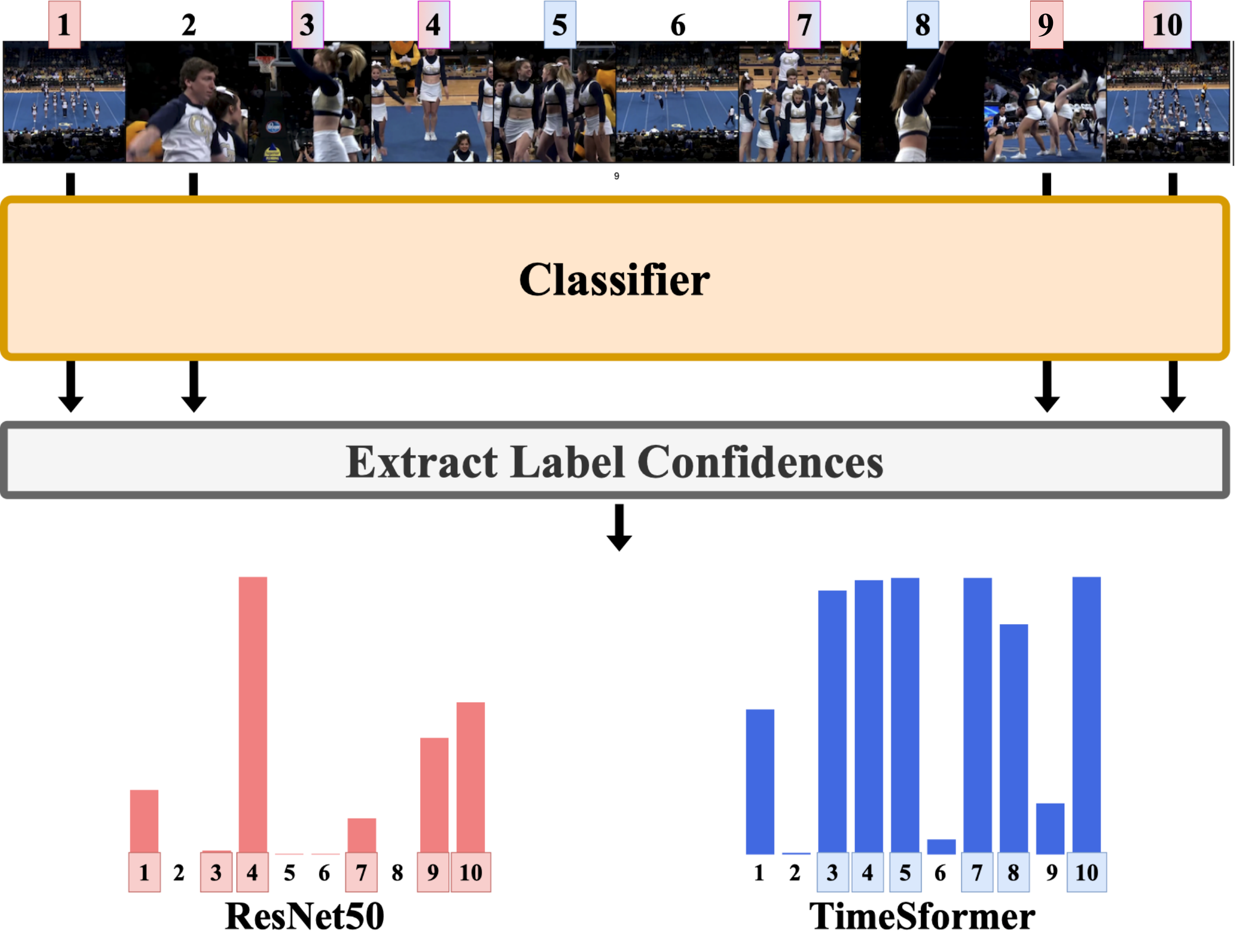}
    \end{minipage}\hfill
    \begin{minipage}{0.48\textwidth}
        \captionof{figure}{\textbf{Illustration of the Semi-Optimal Policy $\bm{\pi_s}$.} We briefly illustrate how $\pi_s$ works on two different architecture. The numbers in the pink and blue boxes represent the frame indices sampled with ResNet50 and TimeSformer as the backbone, respectively, when $N=6$.}
        \label{fig:confidence-based sampler}
    \end{minipage}
\end{figure}

\vspace{0.1cm} \noindent
\textbf{Semi-optimal Policy $\bm{\pi_s}$.}
Under the frame independence condition,
we score the importance of each frame independently.
Specifically, we infer the frame-level confidence scores for all $C$ classes by
treating each frame $\bm{v}_t \in \mathds{R}^{H \times W \times 3}$ as a single-frame video clip $\bm{v}_t \in \mathds{R}^{1 \times H \times W \times 3}$ and take its confidence scores $f_c(\bm{v}_t) \in [0, 1]^{C}$.
We illustrate the mechanism of $\pi_s$ in \cref{fig:confidence-based sampler}.

In order to score the importance of a frame, its confidence scores $f_c(\bm{v}_t)$ on $C$ classes need be aggregated into a single confidence score $c(\bm{v}_t) \in \mathds{R}$.
There are two options for this.
First, we may take the confidence for the true label $y$ as importance score.
That is, $c(\bm{v}_t) = [f_c(\bm{v}_t)]_y$, where $\left[ f_c(\bm{v}_t) \right]_y$ indicates the confidence of frame $\bm{v}_t$ for the label $y$ using the classifier $f_c$.
This approach aims to learn the desired importance score for the actual class.
The second option takes the maximal confidence across all classes $i = 1, ..., C$; that is, $c(\bm{v}_t) = \max_{i=1, ..., C} \left[ f_c(\bm{v}_t) \right]_i$.
Motivation of this approach is to train the sampler with more consistent scores, since the importance may significantly vary depending on the labels even for similar scenes.
Once we have the frame-wise confidence distribution $c(\bm{v}_t)$ for $t = 1, ..., T$, we select the top $N$ frames with highest importance score. That is,
\begin{equation}
    \label{eq:conf-based}
    \pi_s(\bm{v}, y;N) = \operatorname*{top-\textit{N}}_{t = 1, ..., T} c(\bm{v}_t).
\end{equation}

\begin{table}
    \centering
    \begin{minipage}{0.47\textwidth}{
        \centering
        \renewcommand{\tabcolsep}{4pt}
        \resizebox{1.0\linewidth}{!}{
            \begin{tabular}{c|c|rr|cr}
            \toprule
            $f_c$ & Policy & \multicolumn{2}{c|}{A-Net (mAP)} & \multicolumn{2}{c}{M-Kin. (Top-1)} \\
            \midrule
            \multirow{4}{*}{TimeSformer} & $\pi_u$ & 87.0\% & -- & 79.6\% & --\\
            & $\pi_o$ &  91.5\% & +4.5 & 89.3\% & +9.7\\
            & $\pi_s$ &  89.4\% & +2.4 & 84.8\% & +5.2\\
            & All & 89.0\% & +2.0 & 81.2\% & +1.6\\
            \midrule
            \multirow{4}{*}{ResNet50} & $\pi_u$ & 75.3\% & -- & 72.5\% & -- \\
            & $\pi_o$ &  90.5\% & +15.2 & 83.8\% & +11.3 \\
            & $\pi_s$ &  87.4\% & +12.1 & 80.3\% & +7.8 \\
            & All & 77.8\% & +2.5 & 73.6\% & +1.1 \\
            \bottomrule
            \end{tabular}
        }
        \vspace{0.01cm}
        \caption{\textbf{Performance of $\pi_o$ and $\pi_s$ on ActivityNet and Mini-Kinetics}. Relative improvement from $\pi_u$ is provided on the right.}
        \label{tab:oracle_conf_sampler_result}
    }
    \end{minipage}\hfill
    \begin{minipage}{0.51\textwidth}{
        \centering
        \renewcommand{\tabcolsep}{3pt}
        \resizebox{1.0\linewidth}{!}{
            \begin{tabular}{c|c|cccccc}
                \toprule
                \multirow{2}{*}{Dataset} & \multirow{2}{*}{Sampler} & \multicolumn{6}{c}{Sampling Fidelity (\%)} \\ 
                                         &                          & $N=$~1& $N=$~2& $N=$~3& $N=$~4& $N=$~5& $N=$~6    \\ \midrule
                \multirow{3}{*}{A-Net} & Random                 & 10.0  & 20.0  & 30.0  & 40.0  & 50.0 & 60.0  \\ 
                                         & FrameExit                & 10.2 & 19.3 & 29.3 & 39.3 & 49.6 & 59.3 \\ 
                                         & $\pi_s$                      & 100.0    & 74.6 & 73.2 & 75.1 & 78.5 & 81.0 \\ \midrule
                \multirow{3}{*}{M-Kin.} & Random                & 10.0  & 20.0  & 30.0  & 40.0  & 50.0 & 60.0 \\
                                         & FrameExit                & 9.8& 22.5& 32.0 & 42.5 & 51.5 & 62.5 \\ 
                                         & $\pi_s$                     & 100.0    & 61.5 & 65.2 & 70.6  & 71.8  & 80.5  \\ 
                \bottomrule
            \end{tabular}
        }
        \vspace{0.1cm}
        \caption{\textbf{Sampling Fidelity}. 
        Note that we report the expected value of the sampling fidelity for random sampling.
        }
        \label{tab:fidelity}
    }
    \end{minipage}
\end{table}

\vspace{0.1cm} \noindent
\textbf{Empirical Verification.}
We verify that our proposed policy $\pi_s$ reasonably approximates the optimal policy $\pi_o$ in practice.
First, we actually compare the frame sampling performance of $\pi_s$ against $\pi_o$ with two architectures, a CNN and a transformer, on two datasets, ActivityNet-v1.3~\cite{caba2015activitynet} and Mini-Kinetics~\cite{kay2017kinetics}, representing the untrimmed and trimmed video datasets, respectively.
\cref{tab:oracle_conf_sampler_result} compares the performance of a pre-trained classifier when it takes a set of $N=6$ frames selected from $T=10$ frames by various sampling policies:
uniform sampling ($\pi_u$), the optimal policy ($\pi_o$), our semi-optimal policy ($\pi_s$), and using all frames without sampling (All).
The results indicate that $\pi_s$ demonstrates significant improvement over $\pi_u$ and `All' under all settings, most closely approximating the optimal policy, $\pi_o$.

Additionally, we report the sampling fidelity, which indicates the similarity of the sampled set to the optimal set, across various values of $N$. 
Formally, sampling fidelity is defined as $\mathbb{E}_{\bm{v}}\left(|\mathcal{S}_t(\bm{v}) \cap \mathcal{S}_o(\bm{v})| / N \right)$, where $\mathcal{S}_t(\bm{v})$ and $\mathcal{S}_o(\bm{v})$ denote the sampled set of frames from the video $\bm{v}$ by the target policy ($\pi_t$) and $\pi_o$, respectively.
\cref{tab:fidelity} compares sampling fidelity of our $\pi_s$ against two baselines: the random sampling and a deterministic coarse-to-fine sampling policy proposed in FrameExit~\cite{ghodrati2021frameexit}.
We observe that our $\pi_s$ demonstrates significantly higher sampling fidelity than the FrameExit policy, which exhibits fidelity nearly identical to the random policy.
We also provide a qualitative comparison in \cref{sec:success_case}--\ref{sec:failure_case}.

\vspace{-0.1cm}
\subsection{SOSampler: Semi-optimal Policy-based Sampler}
\label{sec:sosampler}

As shown in \cref{subsec:semi}, our semi-optimal policy $\pi_s$ reasonably approximates the optimal policy $\pi_o$.
To train a lightweight network sampler based on this policy,
we propose the Semi-Optimal Sampler (SOSampler), which learns $\pi_s$ instead of $\pi_o$ in a straightforward manner.
\vspace{0.1cm}
\begin{figure}[t]
    \centering
    \includegraphics[width=\textwidth]{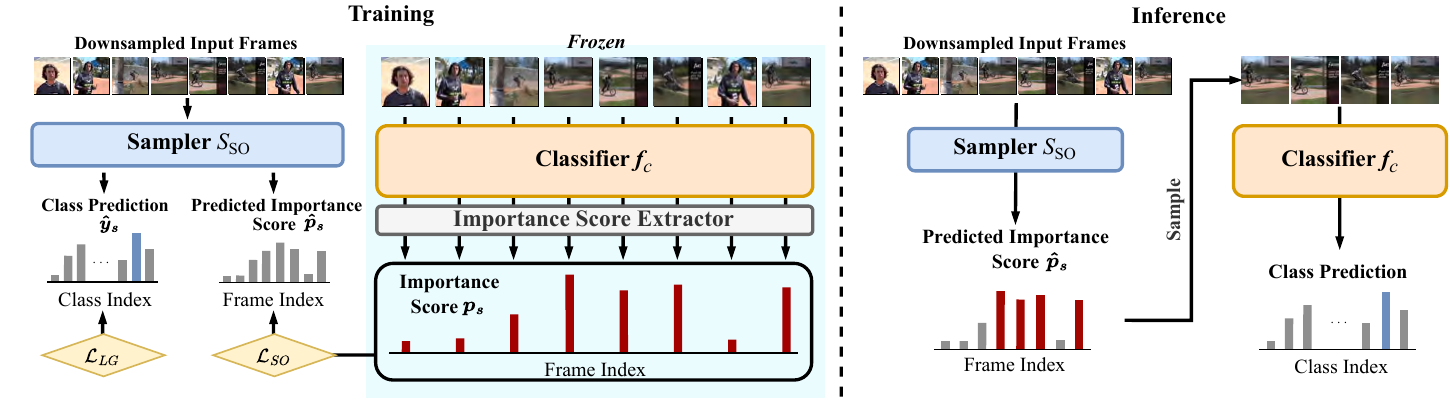}
    \vspace{0pt}
    \vspace{-0.4cm}
    \caption{\textbf{SOSampler Algorithm.} SOSampler consists of a sampler $S_\text{SO}$ and a classifier $f_c$, which can be any model architecture.}
    \label{fig:overview}
\end{figure}
\noindent\textbf{Overview.}
\cref{fig:overview} illustrates an overview of our approach, which consists of a frame sampler $S_\text{SO}$ and an action classifier $f_c$.
At training, we first 
take $T$ candidate frames from the input video and compose a video clip $\bm{v}$.
The classifier $f_c$ takes each candidate frame as a single-frame clip input and conducts classification.
This process is performed individually for each frame, producing $T$ predictions in total.
From these $T$ predictions, we take the representative confidence $c(\bm{v}_t)$ either by taking the true label $y$ or by taking the maximal one, as explained in \cref{subsec:semi}.
We then apply softmax to get a normalized importance distribution $\bm{p}_s \in \mathds{R}^{T}$.
Then, our sampler $S_\text{SO}$ takes the \emph{spatially down-sampled} version of the same $T$ candidate frames.
$S_\text{SO}$ predicts the importance scores $\bm{\hat{p}}_s \in \mathds{R}^{T}$ over them and classifies each frame into the $C$ classes using linear projections.
See \cref{sec:model} for more detailed description.


\vspace{0.1cm} 
\noindent\textbf{Training Objectives.}
We train the sampler $S_\text{SO}$ with two objectives: 1) the semi-optimal policy loss $\mathcal{L}_\text{SO}$, penalizing when the estimated importance score $\bm{\hat{p}}_s$ does not agree with the confidence $\bm{p}_s$ produced by $f_c$, and 2) the label guidance loss $\mathcal{L}_\text{LG}$, penalizing when the predicted class of each frame differs from the true label to ensure that each frame contains information about the video-level class.

For $\mathcal{L}_\text{SO}$, we initially experimented with mean square loss, $\| \bm{\hat{p}}_s - \bm{p}_s \|_2$, but the result was not satisfactory.
Taking inspiration that what we need is a correct ordering of the importance scores, not their exact values, we adopt a pairwise ranking loss~\cite{lee2014lcr} for  $\mathcal{L}_\text{SO}$, which penalizes the model when the relative order of importance between a pair of two frames is reversed:
\vspace{-0.07cm}
\begin{equation}
    \mathcal{L}_\text{SO} = \sum_{(p_i, p_j)\in \Psi}{\text{sign}(p_i - p_j) \cdot \text{max}\left( \gamma+\hat{p}_i - \hat{p}_j, 0\right)},
    \label{eq:ranking}
\end{equation}
where $\Psi = \{(p_i, p_j); p_i > p_j\}$, $\text{sign}(z)$ is the sign function, and $\gamma$ is a hyperparameter indicating a target margin. 
For $\mathcal{L}_\text{LG}$, we use the cross-entropy loss.

The overall loss function is constructed by
\begin{equation}
    \label{eq:total_loss}
    \mathcal{L} = \lambda \cdot \mathcal{L}_\text{SO}(\bm{p}_s,\bm{\hat{p}}_s) + (1-\lambda) \cdot \mathcal{L}_\text{LG}(\bm{\hat{y}}_s, \bm{y}),
\end{equation}
where $\lambda$ is a hyperparameter that adjusts the ratio of the two losses, $\bm{\hat{y}}_s$ denotes the predicted confidence scores by the sampler and $\bm{y}$ is the one-hot encoding of the true label $y$.

\vspace{0.1cm} \noindent
\textbf{Inference.}
At inference, $T$ candidate frames are uniformly sampled from the target video.
As in training, each frame is down-sampled and fed into the sampler $S_\text{SO}$ to obtain the importance scores.
Frames with the top $N$ scores are selected, and those selected frames in the original resolution are fed into $f_c$ to perform the final classification.


\section{Experiment}
\label{sec:exp}

\subsection{Experimental Setup}
\label{sec:exp:setup}

\noindent\textbf{Datasets.}
We evaluate on four public video classification benchmarks:
ActivityNet-v1.3~\cite{caba2015activitynet},
Mini-Kinetics~\cite{kay2017kinetics},
Mini-Sports1M~\cite{karpathy2014sport1m}, and 
COIN~\cite{coin}.
See \cref{sec:dataset} for more detailed description about the datasets.


\vspace{0.1cm}
\noindent\textbf{Experimental Protocol.}
We randomly sample $T$ frames from each video as a training example and uniformly sample $T$ frames from each test video.
We use MobileNetv2TSM~\cite{sandler2018mobilenetv2,lin2019tsm} as the feature extractor of our sampler $S_{\text{SO}}$ and one additional fully-connected layer at the head of the sampler, following~\cite{lin2022ocsampler}.
See \cref{sec:implementation} for more details.

\vspace{0.1cm}
\noindent\textbf{Evaluation Metrics.}
Following the common practice, we measure the mean average precision (mAP), the mean of class average precisions, for ActivityNet, Mini-Sport1M, and COIN datasets.
For Mini-Kinetics, we use top-1 accuracy, which is the ratio of the correctly classified test samples.
For computational cost, we report GFLOPs for all datasets. 
\begin{table}
    \centering
    \begin{minipage}{0.49\textwidth}{
            \centering
            \renewcommand{\tabcolsep}{3pt}
            \resizebox{\linewidth}{!}{
            \begin{tabular}{l|c|@{\hspace{0.3cm}}cccccc}
            \toprule
            \multirow{2}{*}{Methods} & Back- & \multicolumn{2}{c}{ActivityNet} &  \multicolumn{2}{c}{Mini-Kinetics} \\
            & bones & mAP & GFLOPs & Top-1 &  GFLOPs \\
            \midrule
            LiteEval \cite{wu2019liteeval}& & 72.7\% & 95.1 & 61.0\% & 99.0  \\ 
            SCSampler \cite{korbar2019scsampler}& & 72.9\%  & 42.0 & 70.8\% & 41.9 \\
            AR-Net \cite{meng2020arnet}& & 73.8\% & 33.5 & 71.7\% & 32.0 \\
            videoIQ \cite{sun2021videoiq}& Res- & 74.8\% & 28.1 & 72.3\% & 20.4 \\
            AdaFocus \cite{wang2022adafocus}& Net50 & 75.0\% & 26.6 & 72.9\% & 38.6 \\
            FrameExit \cite{ghodrati2021frameexit}& & 76.1\% & 26.1 & 72.8\% & 19.7 \\
            OCSampler \cite{lin2022ocsampler}& & 77.2\% & 25.8 & 73.0\% & {21.6} \\
            \textbf{SOSampler} && {\textbf{77.7\%}}& 25.8 &  {\textbf{73.5\%}}& {21.6} \\
            \midrule
            Ada2D \cite{li2021ada2d} & 
            Slow &
            84.0\% & 701 & 79.2\% & 738 \\
            OCSampler~\cite{lin2022ocsampler} & Only & 87.3\% & 68.2 & 82.6\% & 27.3 \\
            \textbf{SOSampler} & 50 & \textbf{88.0\%} & 64.0 & \textbf{83.0\%} & 27.3 \\
            \midrule
            FrameExit \cite{ghodrati2021frameexit} & \multirow{3}{*}{X3D-S} & 86.0\% & 9.8 & -- & -- \\
            OCSampler~\cite{lin2022ocsampler} &  & 86.6\% & 7.9 & -- & -- \\
            \textbf{SOSampler} &  & {\textbf{87.2\%}} & {7.6} & -- & -- \\
            \midrule
            OCSampler~\cite{lin2022ocsampler} & TimeS- & 83.2\% & {76.8} & 80.7\% & {76.8} \\
            \textbf{SOSampler} & former & \textbf{88.7\%} & 76.8 & 80.7\% & {76.8} \\
            \bottomrule
            
            \end{tabular}
            }
            \vspace{0.2cm}
            \captionof{table}{\textbf{Comparison on ActivityNet-v1.3 and Mini-Kinetics for small $N$ and $T$.} The best performing model is 
            \textbf{bold-faced}.}
            \label{table:sota_anet_k200}
        }
    \end{minipage}\hfill
    \begin{minipage}{0.49\textwidth}{
        \renewcommand{\tabcolsep}{3pt}
        \resizebox{\linewidth}{!}{
        \centering
        \begin{tabular}{l|c|@{\hspace{0.1cm}}cccc}
            \toprule
            \multirow{2}{*}{Methods} & Back- & \multicolumn{2}{c}{Mini-Sports1M} &  \multicolumn{2}{c}{COIN} \\
            & bones & mAP & GFLOPs & Top-1 &  GFLOPs \\
            \midrule        
            LiteEval~\cite{wu2019liteeval}& & 44.7\% & 66.2 & - & - \\
            SCSampler~\cite{korbar2019scsampler} && 44.3\% & 42.0 & 79.8\% & 42.0 \\
            AR-Net~\cite{meng2020arnet} & Res- & 45.0\% & 37.6 & - & - \\
            AdaFuse~\cite{meng2021adafuse} & Net50 &  44.1\% & 60.3 & - & - \\ %
            OCSampler~\cite{lin2022ocsampler} & & 46.7\% & 25.8 & 80.1\% & 25.8 \\
            \textbf{SOSampler} &&{\textbf{48.3\%}} & 25.8 & {\textbf{80.7\%}} & 25.8 \\
            \midrule
            OCSampler \cite{lin2022ocsampler} & TimeS- & 45.6\% & 76.8 & 81.4\% & 76.8  \\
            \textbf{SOSampler} & former & \textbf{49.1\%} & 76.8 & \textbf{87.7\%} & 76.8 \\
            \bottomrule
            \end{tabular}
        }
        \vspace{0.2cm}
        \captionof{table}{\textbf{Comparison on Mini-Sports1M and COIN Comparison on for small $N$ and $T$}. The best performing model is \textbf{bold-faced}.}
        \label{table:sota_sport_coin}
    }
    \end{minipage}
\end{table}
\subsection{Results and Analysis}
\label{sec:exp:results}
\vspace{0.1cm}

\noindent \textbf{Comparison with Small $N$ and $T$.}
To compare with existing methods under the same conditions,\footnote{Results for OCSampler with ResNet50 in \cref{table:sota_anet_k200} is our reproduction using pre-trained weights and settings provided by the original author. They slightly lag behind on Mini-Kinetics, by 0.7\%.} we first evaluate with the common setting of $T = 10$ and $N = 6$ (except for Mini-Kinetics with ResNet50, where $N = 5$).

%

As seen in \cref{table:sota_anet_k200}--\ref{table:sota_sport_coin}, our approach consistently outperforms previous methods across all four datasets and backbone architectures, highlighting the effectiveness of learning $\pi_s$ even when the search space is not very large.
Nevertheless, comparing \cref{table:sota_anet_k200} to \cref{tab:oracle_conf_sampler_result}, we observe that SOSampler still does not fully emulate $\pi_s$, possibly due to the loss when it transfers the label information.
A future work may further address this issue to fill the gap.

\begin{figure}
    \begin{minipage}[b]{0.56\linewidth}
        \centering
            \centering
            \renewcommand{\tabcolsep}{3pt}
            \resizebox{\linewidth}{!}{
            \begin{tabular}{@{}l|l|l|ccc@{}}
            \toprule
            \multirow{2}{*}{Dataset}     & \multirow{2}{*}{Backbone} & \multirow{2}{*}{Method} & \multicolumn{3}{c}{$N$~/~$T$} \\
                                         &                           &         & 8~/~30          & 16~/~60         & 32~/~100         \\ 
            \midrule
            \multirow{6}{*}{ActivityNet} & \multirow{3}{*}{ResNet50}      & Uniform           & 77.1\%       & 79.4\%       & 80.4\%       \\
                                         &                              & OCSampler         & 78.0\%        & 79.1\%        & 80.1\%        \\
                                         &                              & \textbf{SOSampler}         & \textbf{78.7\%}        & \textbf{80.2\%}        & \textbf{81.1\%}        \\ 
            \cmidrule(l){2-6}
                                         & \multirow{3}{*}{TimeSformer}  & Uniform           & 88.5\%        & 89.9\%        & 90.3\%        \\
                                         &                              & OCSampler         & 85.0\%        & 85.3\%        & 84.5\%        \\
                                         &                              & \textbf{SOSampler}         & \textbf{89.5\%}        & \textbf{90.1\%}        & \textbf{90.5\%}        \\ 
            \midrule
            \multirow{6}{*}{Mini-Sports1M} & \multirow{3}{*}{ResNet50} & Uniform  & 46.9\% & 48.8\% & 49.1\% \\
                                         &  & OCSampler  & 48.6\% & 49.6\% & 50.0\% \\
                                         &  & \textbf{SOSampler}  & \textbf{50.0\%} & \textbf{51.1\%} & \textbf{51.5\%} \\
            \cmidrule(l){2-6}
                                         & \multirow{3}{*}{TimeSformer} & Uniform  & 53.9\% & 55.6\% & 56.8\% \\
                                         &  & OCSampler & 48.9\% & 49.6\% & 50.0\% \\
                                         &  & \textbf{SOSampler} & \textbf{55.1\%} & \textbf{56.9\%} & \textbf{57.8\%} \\
            \bottomrule
            \end{tabular}
            }
            \vspace{0.3cm}
            \captionof{table}{\textbf{Experiment on long videos for large $N$ and $T$}. The best performing model is \textbf{bold-faced}.}
            \label{tab:large_tn}
    \end{minipage}
    \hfill
    \begin{minipage}[b]{0.41\linewidth}
        \centering
        \includegraphics[width=\textwidth]{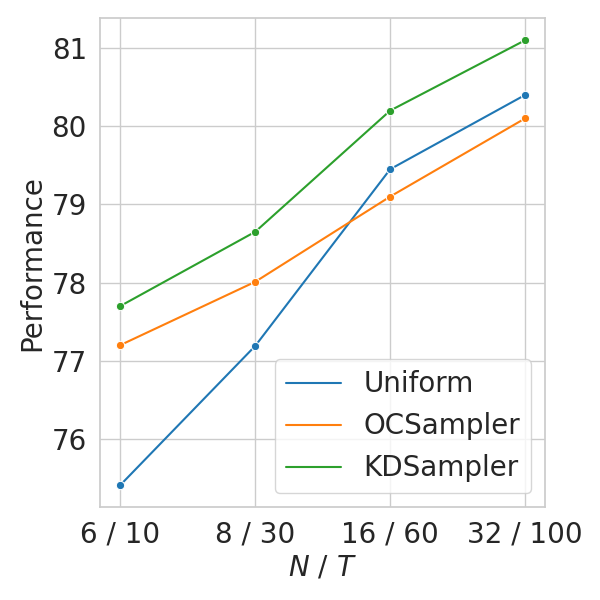}
        \caption{\textbf{Performance by sampler across N/T on ActivityNet-v1.3.}}
        \label{fig:anet_tn_graph}
    \end{minipage}
\end{figure}

\vspace{0.1cm}
\noindent\textbf{Comparison with Large $N$ and $T$.}
We conduct experiments on large $N$ and $T$ to verify the effect of reduced search space by our proposed policy $\pi_s$, comparing with OCSampler~\cite{lin2022ocsampler} and uniform sampling as baselines.
Note that OCSampler and SOSampler have exactly the same architecture -- the only difference lies in the learning objective.
We evaluate on the ActivityNet-v1.3 and Mini-Sports1M datasets, which consist of sufficiently long videos\footnote{Results for short videos are discussed in~\cref{appendix:large_nt_short}.}, to ensure the frame independence even for a large $T$.

As shown in \cref{tab:large_tn}, our approach consistently achieves competitive performance regardless of the backbone architectures and with various values of $N$ and $T$.
With the OCSampler, on the other hand, the performance improvement is inconsistent and it sometimes underperforms even than the uniform sampling.
This is illustrated in \cref{fig:anet_tn_graph} as well; the performance of OCSampler lags behind the uniform sampling from $N = 16, T = 60$ and beyond, indicating that directly searching the $O(T^N)$ space is not scalable for large $N$ and $T$.
Our semi-optimal policy does not suffer from this, recalling that our approach adopts the same architecture as OCSampler.
Additionally, we observe that OCSampler does not perform well with the TimeSformer backbone.
We also attribute this to the more complex search space of transformer-based classifiers compared to CNN-based ones.
Based on the results, we conclude that our approach to reduce the complexity of search space is indeed effective for large $N$ and $T$, which has been challenging for existing methods.


\vspace{0.1cm}
\noindent\textbf{Qualitative Analysis.}
In \cref{sec:success_case}--\ref{sec:failure_case}, we illustrate success and failure cases of our SOSampler, comparing with sampling policies of $\pi_o$ and $\pi_s$.

\subsection{Ablation Study}
\label{sec:exp:ablation}

\noindent\textbf{Loss function.}
\cref{tab:loss_ablation} reports the result of an ablation study on the loss functions, conducted on ActivityNet-v1.3.
`MSE' and `Ranking' indicate the two choices for $\mathcal{L}_\text{SO}$ presented in \cref{sec:sosampler}, while `Label' and `Max' indicate how we obtain the importance score of each frame $c(\bm{v}_t)$ from the confidence distribution $f_c(\bm{v}_t)$ over all classes.


As we mentioned earlier, the ranking loss in Eq.~\eqref{eq:ranking} outperforms the MSE loss for $\mathcal{L}_\text{SO}$.
We also observe that the max-aggregation leads to stronger performance than the label confidence, probably because it learns consistent scores across features.
The label guidance loss $\mathcal{L}_\text{LG}$ turns out to slightly improve the overall performance.

\begin{table}
    \centering
    \begin{minipage}{0.38\textwidth}
        \centering
        \renewcommand{\tabcolsep}{4pt}
        \resizebox{\linewidth}{!}{
            \begin{tabular}{cccc|c|c}
                \toprule
                \multicolumn{4}{c|}{${\mathcal{L}_{\text{SO}}}$} & \multirow{2}{*}{${\mathcal{L}_{\text{LG}}}$} & \multirow{2}{*}{mAP (\%)} \\
                MSE & Ranking & Label & Max & & \\
                \midrule
                \checkmark & & \checkmark & & & 87.38 \\
                \checkmark & & \checkmark & & \checkmark & 87.51 \\
                & \checkmark & \checkmark & & & 88.21 \\
                & \checkmark & \checkmark & & \checkmark & 88.40 \\
                & \checkmark & & \checkmark & \checkmark & \textbf{88.65} \\
                \bottomrule
            \end{tabular}
        }
        \vspace{0.2cm}
        \captionof{table}{\textbf{Ablation study on losses.}}
        \label{tab:loss_ablation}
    \end{minipage}\hfill
    \begin{minipage}{0.58\textwidth}
        \renewcommand{\tabcolsep}{3pt}
        \resizebox{\linewidth}{!}{
            \begin{tabular}{c|cccc|cccc}
                \toprule
                $T$ & \multicolumn{4}{c|}{$10$} & $6$ & $10$ & $16$ & $24$\\
                \midrule
                $N$ & $2$ & $4$ & $6$ & $8$ & \multicolumn{4}{c}{$6$}\\
                \midrule
                mAP & 71.1\% & 76.2\% & 77.7\% & \bf{77.9\%}  & 75.3\% & 77.7\% & 78.2\% & \bf{78.4\%} \\ 
                GFLOPs & \bf{9.3} & 17.6 & 25.8 & 34.1   & \bf{24.7} & 25.8 & 26.4 & 27.2 \\
                \bottomrule
            \end{tabular}
        }
        \vspace{0.2cm}
        \captionof{table}{\textbf{Ablation study on $T$ and $N$.}}
        \label{tab:t_ablation}
    \end{minipage}
\end{table}

\vspace{0.1cm}
\noindent\textbf{Exploration on $N$ and $T$.}
We fix $N = 6$ and $T = 10$ in \cref{table:sota_anet_k200}--\ref{table:sota_sport_coin} mainly for the purpose of comparison with previous models.
We further explore various combinations of $N \in \{2, 4, 6, 8\}$ and $T \in \{6, 10, 16, 24\}$ to better understand our method.
As seen in \cref{tab:t_ablation}, performance consistently improves with a larger $N$, but the gain diminishes beyond $N \ge 6$.
Also in the case of $N=6$, we observe consistently improved performance with larger $T$.

\section{Conclusion}
\label{sec:summary}




In this paper, we address the scalability challenge of the frame sampling task, proposing our novel novel semi-optimal policy $\pi_s$ to dramatically reduce the search space itself from $O(T^N)$ to $O(T)$, supported by empirical evidence of frame independence.
Through extensive experiments, we demonstrate that $\pi_s$ effectively approximates the optimal policy.
Furthermore, across all datasets and architectures, the sampler which learns $\pi_s$ instead of $\pi_o$ outperforms existing methods both for small and large $N$ and $T$.
These results suggest that our new approach, which changes the search space itself rather than the exploration method, is more effective than previous approaches.
However, the frame independence assumption we propose does not hold in all scenarios, and there remains potential for further performance improvements in the sampler learning our proposed $\pi_s$. Addressing these limitations will be an important direction for future work.

\section*{Acknowledgements}

\noindent
This work was supported by the New Faculty Startup Fund from Seoul National University, by Samsung Electronics Co., Ltd (IO230414-05943-01, RAJ0123ZZ-80SD), by Youlchon Foundation (Nongshim Corp.), and by National Research Foundation (NRF) grants (No.
2021H1D3A2A03038607/50\%, RS-2024-00336576/10\%, RS-
2023-00222663/5\%) and Institute for Information \& communication Technology Planning \& evaluation (IITP) grants (No. RS-2024-00353131/25\%, RS-2022-II220264/10\%), funded by the government of Korea.

\bibliography{egbib}
\clearpage

\appendix

\pagenumbering{roman}
\renewcommand\thetable{\Roman{table}}
\renewcommand\thefigure{\Roman{figure}}
\setcounter{table}{0}
\setcounter{figure}{0}

\setcounter{page}{1}

\section*{Appendix}

\section{Detailed Descriptions}
\label{sec:detailed}
\subsection{Model Configuration}
\label{sec:model}

\noindent\textbf{Sampler.}
The sampler $S_\text{SO}$ takes a sequence of $T$ frames as input, composed of a lightweight feature extractor $f_s$, an importance predictor $h_s$, and an action classifier $h_c$.

Given $T$ candidate frames $\bm{v} \in \mathds{R}^{T \times 3 \times H \times W}$, our sampler first spatially downsamples them to $\bm{v'} \in \mathds{R}^{T \times 3 \times H' \times W'}$, where $W' < W$ and $H' < H$.
Then, a 2D image representation network $f_s: \mathds{R}^{3 \times H' \times W'} \rightarrow \mathds{R}^{D}$ extracts frame-level features, where $D$ denotes the dimensionality of the feature.
Inferring all $T$ frames using $f_s$, a feature map
\begin{equation}
\bm{z} =
\{f_s(\bm{v}'_1), ..., f_s(\bm{v}'_T)\} \in \mathds{R}^{T \times D}
\end{equation}
is constructed, where $\bm{v}'_i$ denotes the $i$-th frame of $\bm{v}'$.

Then, our sampler conducts two downstream tasks using the extracted features $\bm{z}$.
First, it estimates the frame importance score $\bm{p}_s$ using a regressor $h_s: \mathds{R}^{T \times D} \rightarrow \mathds{R}^{T}$:
\begin{equation}
\label{eq:importance_prediction}
\bm{\hat{p}}_s = \{h_s(\bm{z}_1), ..., h_s(\bm{z}_T)\}.
\end{equation}
Second, it performs the downstream classification task.
Using a frame-level classifier $h_c: \mathds{R}^{T \times D} \rightarrow \mathds{R}^{C}$, it predicts the relevance of each frame $t = 1, ..., T$ for the $C$ classes, and these predictions are aggregated over the $T$ frames to a video-level prediction by taking the average:
\begin{equation}
\bm{\hat{y}}_s = \frac{1}{T} \left( h_c(\bm{z}_1) + ... + h_c(\bm{z}_T) \right).
\end{equation}
Note that $h_c$ is used only during training to make the backbone $f_s$ learn the label information.
We simply implement $h_s$ and $h_c$ with linear projections followed by a softmax.

\vspace{0.1cm}
\noindent\textbf{Classifier.}
The classifier $f_c$ can be any visual recognition model, such as a 2D or 3D CNN, or a Transformer, pretrained and frozen throughout the training.
During training, $f_c$ is used to compute the importance score, which serves as a pseudo-label to train the sampler $S_\text{SO}$ by distilling the knowledge from the classifier $f_c$.

During inference, $f_c$ is used to perform the downstream task on a clip $\bm{v}^s \in \mathds{R}^{N \times 3 \times H \times W}$ of sampled $N$ frames:
\begin{equation}
\bm{\hat{y}} = f_c(\bm{v}^s).
\end{equation}
Our goal is to train the sampler $S_\text{SO}$ so that the frozen classifier $f_c$ predicts $\bm{\hat{y}}$ close to the ground truth label $\bm{y}$, the one-hot encoding of the true label $y$.

\subsection{Dataset}
\label{sec:dataset}
ActivityNet-v1.3 includes 10,024 training and 4,926 validation videos. The average video length is 117 seconds with an average of 3,335 frames, covering 200 categories.
Mini-Kinetics, a subset of Kinetics400, comprises 121,215 training and 9,867 validation videos. The videos have an average duration of 10 seconds with an average of 261 frames, covering 200 categories.
Mini-Sports1M, a subset of Sports1M~\cite{gao2020listentolook}, consists of 14,586 training and 4,855 validation videos. The average video length is 330 seconds with an average of 4,467 frames, covering 487 action classes.
COIN is composed of 11,827 YouTube videos related to 180 different tasks. The videos have an average length of 141 seconds with an average of 4,009 frames.

\subsection{Implementation Details}
\label{sec:implementation}
We follow the preprocessing steps outlined in \cite{lin2022ocsampler}. We sample $T$ frames from each video as a training example. All frames are randomly scaled and cropped to $224 \times 224$, followed by random flipping for augmentation. We then reduce the resolution of each frame to $128 \times 128$ before feeding them into our sampler $S_\text{SO}$.
During inference, we uniformly sample $T$ frames from a test video, resize them to $128\times128$, and feed them to the sampler $S_\text{SO}$.
Then, we feed the original $224 \times 224$ images of the selected frames to $f_c$.

For the pretrained classifier weights, we utilize the pretrained weights provided by \cite{ghodrati2021frameexit} on the ActivityNet-v1.3 and Mini-Kinetics datasets with the ResNet50 classifier. For other datasets and architectures, we train the classifier from scratch.

To train our SOSampler, we use a learning rate of $10^{-3}$ and set $\lambda = 0.99$ for all datasets. We optimize our loss function in Eq.~\eqref{eq:total_loss} using the stochastic gradient descent (SGD) optimizer with a momentum of 0.9 and weight decay set to $10^{-4}$. We employ cosine annealing as a learning rate scheduler without warm-up.

We implement our method using PyTorch and train on a single NVIDIA A100 GPU with 40GB of memory.

\section{Additional Results}
\label{sec:additional}

\begin{table}[t]
    \centering
    \label{tab:tn_ablation}
    \resizebox{0.8\linewidth}{!}{
    \begin{tabular}{@{}l|l|l|ccc@{}}
    \toprule
    \multirow{2}{*}{Dataset}     & \multirow{2}{*}{Backbone} & \multirow{2}{*}{Method} & \multicolumn{3}{c}{$N$~/~$T$} \\
                                 &                           &                   & 8~/~30          & 16~/~60   & 32~/~100         \\ 
    \midrule
    \multirow{4}{*}{Mini-Kinetics} & \multirow{2}{*}{ResNet} & OCSampler   & 73.52\%  & 74.00\%  & 74.17\%  \\
                                 &                              & SOSampler   & \textbf{73.79\%}  & \textbf{74.46\%}  & \textbf{74.65\%}  \\
    \cmidrule(l){2-6}
                                 & \multirow{2}{*}{TimeSformer} & OCSampler        & 79.13\%  & 78.05\%  & 76.33\%  \\
                                 &                              & SOSampler         & \textbf{79.93\%}  & \textbf{80.44\%}  & \textbf{81.32\%}  \\ 
    \midrule
    \multirow{4}{*}{COIN} & \multirow{2}{*}{ResNet} & OCSampler  & 78.69\% & 79.79\% & 80.06\% \\
                         &  & SOSampler  & \textbf{79.13\%} & \textbf{80.21\%} & \textbf{81.02\%} \\
    \cmidrule(l){2-6}
     & \multirow{2}{*}{TimeSformer} & OCSampler  & 80.88\% & 80.90\% & 81.20\% \\
                                 &  & SOSampler  & \textbf{86.52\%} & \textbf{87.37\%} & \textbf{88.08\%} \\
    \bottomrule
    \end{tabular}
    }
    \vspace{0.2cm}
    \captionof{table}{\textbf{Experiment on short video datasets for large $N$ and $T$}. The best performing model is \textbf{bold-faced}.}
    \label{tab:short_video}
\end{table}

\subsection{Comparison with Large $N$ and $T$ on Short Videos}
\label{appendix:large_nt_short}

In short videos, as $T$ increases, the FPS becomes significantly higher, weakening our assumption of independence between frames. Therefore, our approach does not sufficiently improve the performance as $N$ and $T$ increase. However, it consistently shows an upward trend and still outperforms OCSampler in all settings. This result suggests that our method is still superior to the existing method, even in the FPS range where our assumption of independence between frames is weak.

\subsection{Computational Efficiency}
While GFLOPs serve as a metric for measuring the efficiency of a model, it does not provide the actual running time.
Therefore, we additionally compare the actual inference time, namely throughput, by measuring the video processing speed per second.
The experiments are conducted using ResNet50 on the ActivityNet-v1.3 dataset, and all experiments are performed on a single NVIDIA Xp GPU.
As seen in \cref{tab:throughput}, we demonstrate improved accuracy of our proposed method (SOSampler) over existing methods, achieving a reduction of approximately 16.3\% in GFLOPs and a 15\% enhancement in throughput.

\begin{table}[h]
    \centering
    \footnotesize
        \begin{tabular}{l|@{\hspace{0.4cm}}cccc}
        \toprule
        \multirow{2}{*}{Methods} & \multirow{2}{*}{mAP} & \multirow{2}{*}{GFLOPs} &  \multicolumn{2}{c}{Throughput}\\
        &&& (Videos/s) \\
        \midrule
        AdaFrame~\cite{wu2019adaframe} & 71.5\% & 79.0 & 6.4 \\
        FrameExit~\cite{ghodrati2021frameexit} & 76.1\% & 26.1 & 19.1 \\
        AR-Net~\cite{meng2020arnet} & 73.8\% & 33.4 & 23.1 \\
        AdaFocus~\cite{wang2022adafocus} & 75.0\% & 26.6 & 44.9 \\
        OCSampler~\cite{lin2022ocsampler} & 77.2\% & 25.8 & 107.7 \\
        \textbf{SOSampler} (ours) &\textbf{77.3\%} & \textbf{21.6} & \textbf{123.9} \\
        \bottomrule
        \end{tabular}
    \vspace{0.2cm}
    \caption{\textbf{Comparison of computational overhead} in GFLOPs and throughput. (ResNet50 on ActivityNet)}
    \label{tab:throughput}
\end{table}
\begin{figure}
  \includegraphics[width=6.5cm]{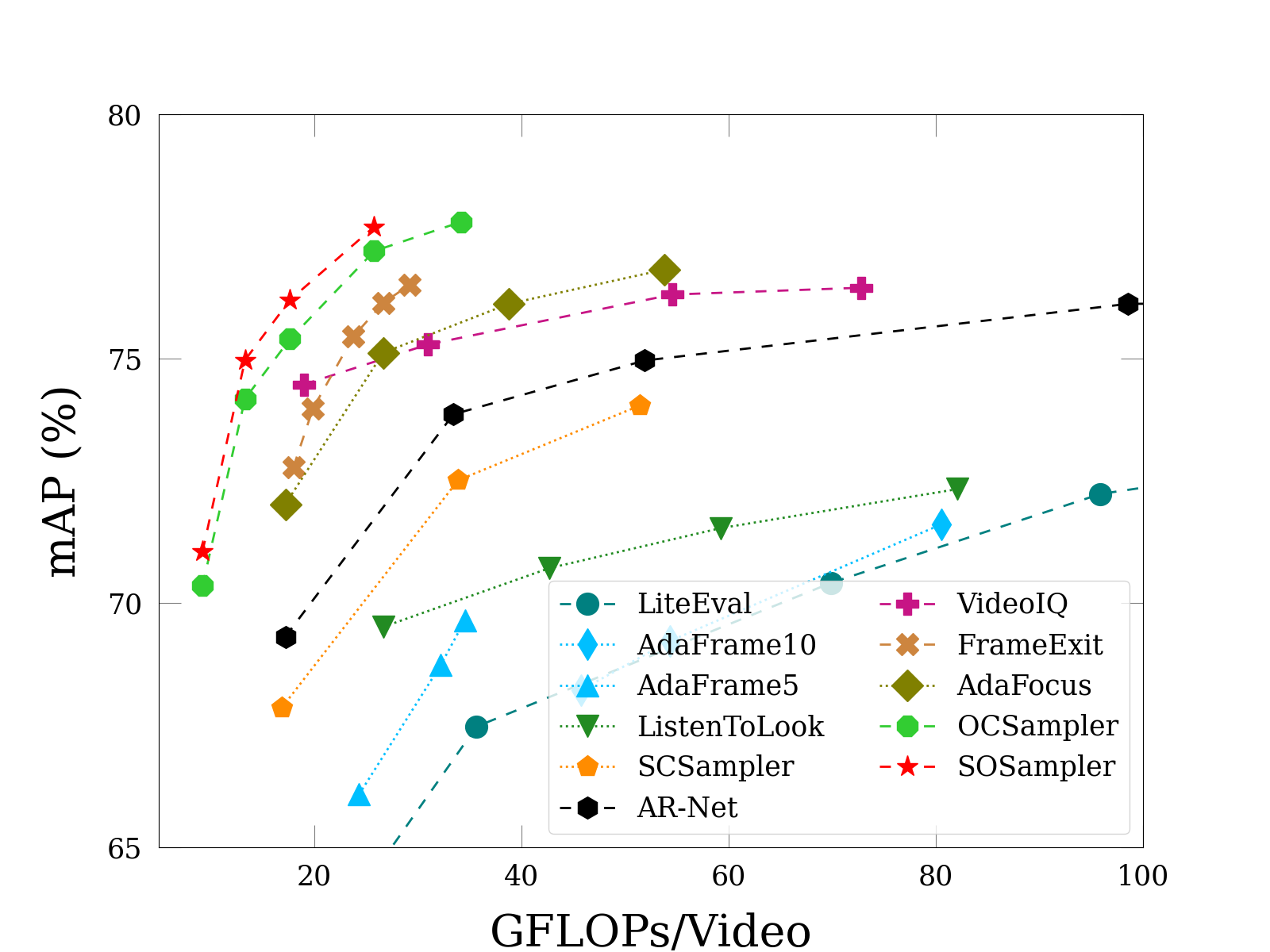}
  \includegraphics[width=6.5cm]{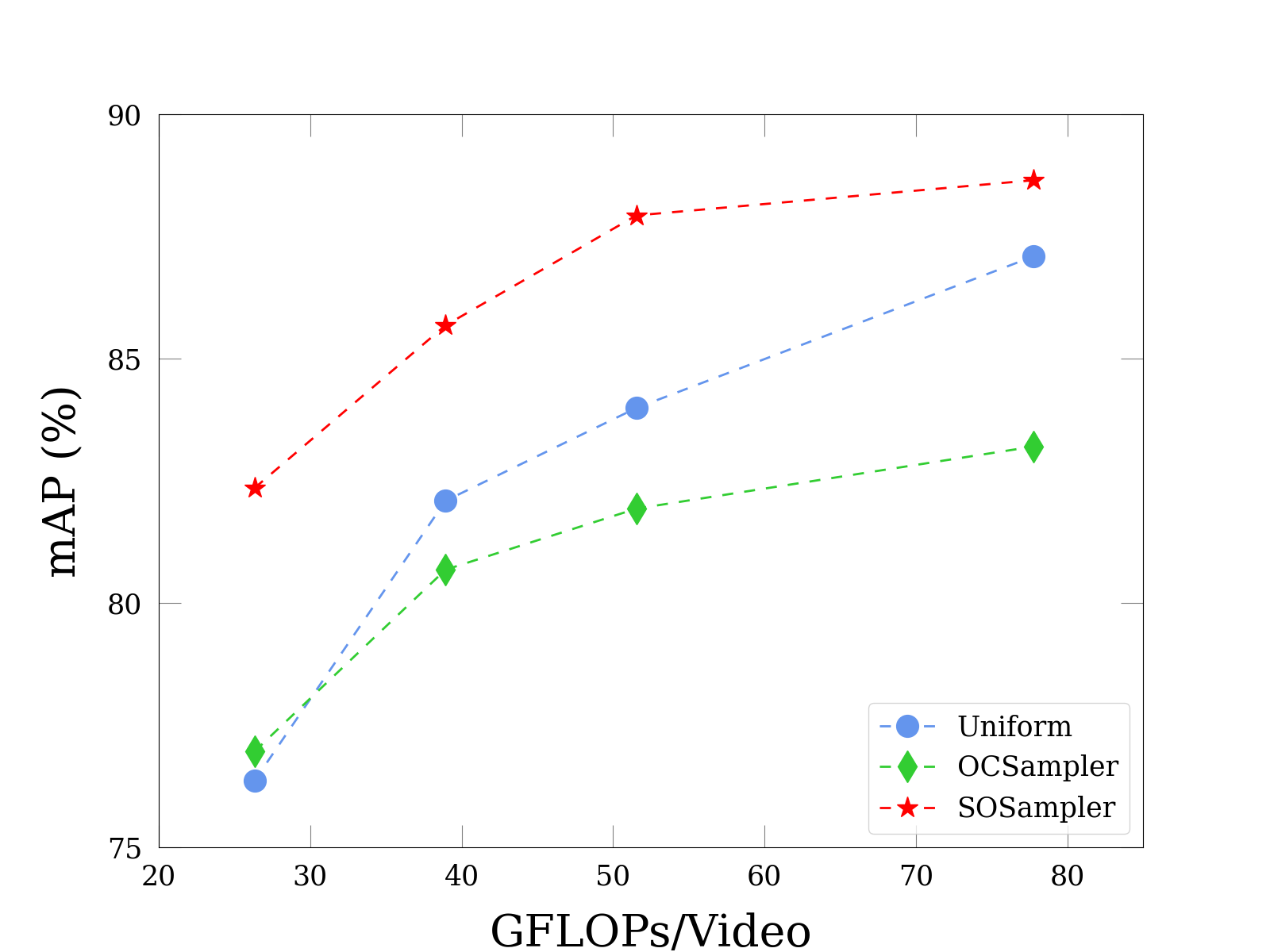}
  \vspace{0.2cm}
  \caption{\textbf{Mean Average Precision (\%) vs. efficiency (GFLOPs) on ActivityNet.} With a ResNet classifier (\emph{left}), OCSampler~\cite{lin2022ocsampler} is the second best after ours. With TimeSformer in (\emph{right}), however, it even underperforms than the uniform sampling. On the other hand, our approach outperforms all baselines with both ResNet50 and TimeSformer.}
  \label{fig:acc_vs_eff}
\end{figure}

\subsection{Performance and Efficiency Curve}
In \cref{fig:acc_vs_eff}(left), we compare our approach to existing methods with varying computational costs, with a varied number of sampled frames $N={2, 3, 4, 6}$ on a ResNet50~\cite{he2016resnet} classifier.
Our method leads all other compared methods, using significantly lower computational cost than most baseline methods, showing marginal improvement over OCSampler~\cite{lin2022ocsampler}.

We additionally conduct a performance and efficiency comparison using the TimeSformer~\cite{bertasius2021timesformer} backbone.
The experiment, like the one performed on ResNet50, measures the changes in computational cost for $N={2, 3, 4, 6}$ and the comparison is exclusively with OCSampler, previously the highest-performing model.
As shown in \cref{fig:acc_vs_eff}(right), within the TimeSformer architecture, our model significantly improves the performance over OCSampler.

\section{Sampling Cases}
\label{sec:sampling_cases}

In \cref{subsec:semi}, we introduce $\pi_s$ and demonstrate that it approximates $\pi_o$ through \cref{tab:oracle_conf_sampler_result} and \cref{tab:fidelity}.
By showing that SOSampler, which learns $\pi_o$ instead of $\pi_s$, outperforms existing methods across various datasets and architectures, we demonstrate the effectiveness of $\pi_s$.

In this section, for a better understanding of our approach, we visually illustrate multiple examples showing that SOSampler successfully approximates $\pi_s$ as well as some cases where it does not.

\subsection{Illustration of Successful Sampling}
\label{sec:success_case}

\begin{figure*}[h]
    \centering 
    \includegraphics[width=1.0\textwidth]{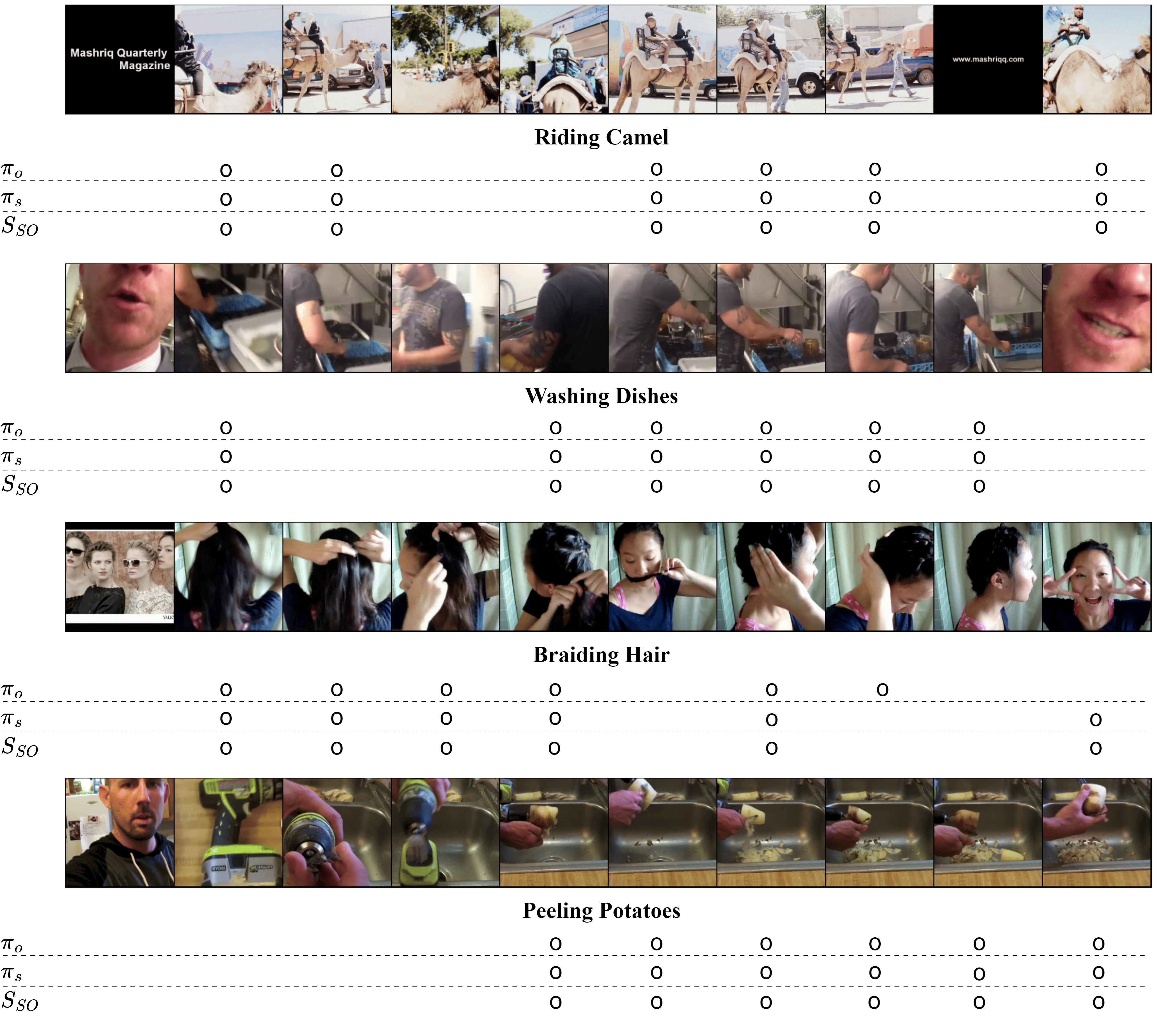} 
    \vspace{0.2cm}
    \caption{\textbf{Sampling policy comparison with $\pi_s$, $\pi_o$ for success cases of SOSampler.}
    }
    \label{fig:success}
\end{figure*}
In \cref{fig:success}, we present qualitative examples of successful sampling by SOSampler, comparing the results with those of $\pi_o$ and $\pi_s$.
For the ``Riding Camel'' example, although the 4th and 5th frames feature a camel, they are not selected due to the lack of clear information about riding compared to other scenes.
In the ``Braiding Hair'' example, $\pi_o$ and $\pi_s$ choose slightly different frames, with SOSampler following the selection pattern of $\pi_s$.
In the case of ``Peeling Potatoes'', it is observed that all policies effectively sample only the portions where potatoes appear.

These results demonstrate that $\pi_o$ and $\pi_s$ possess similar policies. Additionally, they indicate that SOSampler can effectively learn the policy of $\pi_s$.

\subsection{Failure Cases}
\label{sec:failure_case}

\begin{figure*}[h]
    \centering 
    \includegraphics[width=1.0\textwidth]{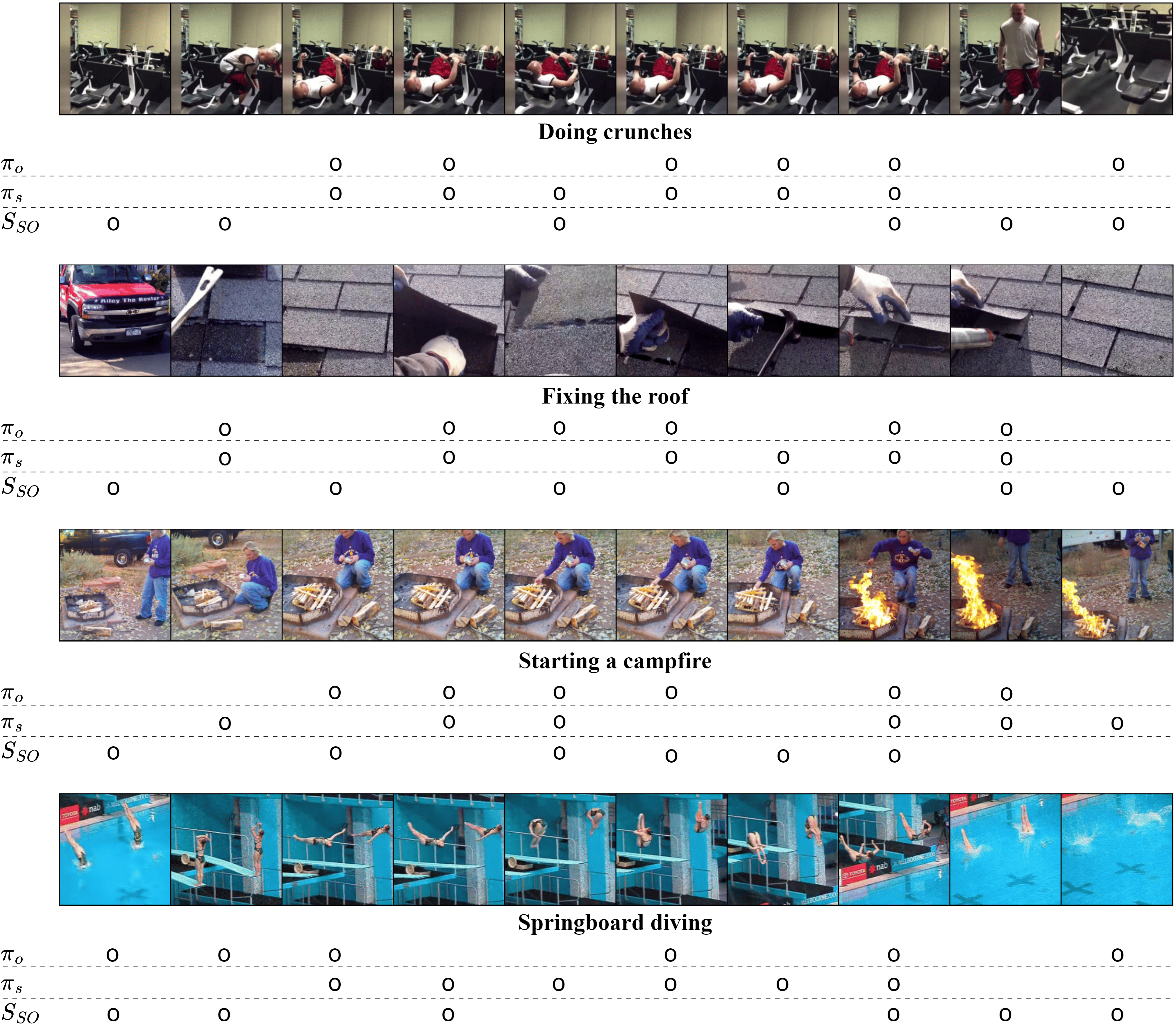} 
    \vspace{0.2cm}
    \caption{\textbf{Sampling policy comparison with $\pi_s$, $\pi_o$ for failure cases of SOSampler.}
    }
    \label{fig:fail}
\end{figure*}

As shown in \cref{fig:success}, in most cases, $S_\text{SO}$ demonstrates a sampling policy similar to $\pi_s$.
In \cref{fig:fail}, however, we showcase a few scenarios where they significantly differ.
In the case of the ``Doing Crunches'', $\pi_s$ effectively samples the segments where a man is performing crunches, while $S_\text{SO}$ samples scattered frames throughout the video.
For the ``Fixing the Roof'', $\pi_s$ appropriately selects scenes of repairing damaged roofs, while $S_\text{SO}$ chooses unrelated frames as well.
In the case of ``Starting a Campfire'', $S_\text{SO}$ seems to summarize the video well, but the sampling policy of $\pi_o$ indicates that the classifier $f_c$ prefers the scenes of installing firewood and starting the fire.
Interestingly, in the ``Springboard Diving'' example, $S_\text{SO}$ even appears to better emulate $\pi_o$ than $\pi_s$ does.

\end{document}